%% file: main_full.tex
\documentclass{article} 
\usepackage{collas2026_conference,times}
\usepackage{easyReview}

\input{math_commands.tex}

\usepackage{hyperref}
\hypersetup{
    colorlinks=true,
    linkcolor=red,
    filecolor=magenta,
    urlcolor=blue,
    citecolor=purple,
    pdftitle={Overleaf Example},
    pdfpagemode=FullScreen,
    }

\usepackage{url}            
\usepackage{booktabs}       
\usepackage{amsfonts}       
\usepackage{nicefrac}       
\usepackage{microtype}      
\usepackage[dvipsnames]{xcolor}
\usepackage{graphicx}
\usepackage{pifont}
\usepackage{multirow} 
\usepackage{subcaption}
\usepackage{algorithm}
\usepackage{algorithmic}
\usepackage{amsmath}
\usepackage{amssymb}
\usepackage{xspace}
\usepackage{makecell}
\usepackage{xcolor}
\definecolor{datablue}{RGB}{0, 102, 204} 

\newcommand{\jumplora}{\textsc{JumpLoRA}}

\title{JumpLoRA: Sparse Adapters for Continual Learning\\ in Large Language Models}

\author{Alexandra Dragomir$^{1,*}$\\
Bitdefender
\And
Ioana Pintilie$^{1,*}$\\
Bitdefender
\And
Antonio Barbalau$^1$\\
Bitdefender
\And
Marius Dragoi$^1$\\
Bitdefender
\And
Florin Brad$^1$\\
Bitdefender
\And
Cristian Daniel Paduraru$^1$\\
Bitdefender
\And
Alexandru Tifrea$^2$\\
ETH Zurich
\And
Elena Burceanu$^1$\\
Bitdefender
\And
Radu Tudor Ionescu$^{3,}$\thanks{Corresponding e-mail(s): \texttt{\{aledragomir, ipintilie\}@bitdefender.com; raducu.ionescu@gmail.com}} \\
University of Bucharest
}

\preprintcopy 

\begin{document}

\maketitle

\begin{abstract}
Adapter-based methods have become a cost-effective approach to continual learning (CL) for Large Language Models (LLMs), by sequentially learning a low-rank update matrix for each task. To mitigate catastrophic forgetting, state-of-the-art approaches impose constraints on new adapters with respect to the previous ones, by targeting either subspace or coordinate-wise interference. 
In this paper, we propose \jumplora, a novel framework to adaptively induce sparsity in the Low-Rank Adaptation (LoRA) blocks through the use of JumpReLU gating. The method achieves dynamic parameter isolation, which helps prevent task interference. We demonstrate that our method is highly modular and compatible with LoRA-based CL approaches. Specifically, it significantly boosts the performance of IncLoRA and outperforms the leading state-of-the-art CL method, ELLA. To reproduce the reported results, we publicly release our code at \url{https://github.com/alexandra-dragomir/JumpLoRA}.
\end{abstract}


\section{Introduction}
The rapid advancement of Large Language Models (LLMs) has revolutionized natural language processing \citep{DBLP:conf/nips/VaswaniSPUJGKP17, DBLP:conf/nips/BrownMRSKDNSSAA20}, enabling unprecedented performance across diverse generative and reasoning tasks \citep{DBLP:journals/corr/abs-2303-18223, DBLP:journals/corr/abs-2402-06196}. However, LLMs are typically static once trained, without an embedded ability to integrate new data without undergoing costly retraining. Continual Learning (CL) \citep{DBLP:journals/corr/abs-2402-01364, DBLP:journals/csur/ShiXWQWWWEW26, DBLP:journals/corr/abs-2603-12658} aims to address this by allowing models to acquire new knowledge from a sequential stream of tasks. 

One fundamental challenge in CL is mitigating catastrophic forgetting \citep{mccloskey1989catastrophic, french1999catastrophic,DBLP:journals/corr/KirkpatrickPRVD16}, where the sequential acquisition of new information results in the abrupt loss of previously learned knowledge. This phenomenon is inherent to the stability-plasticity trade-off \citep{DBLP:journals/cogsci/Grossberg87,abraham2005memory,DBLP:journals/nature/DohareHLRMS24, DBLP:journals/pami/LangeAMPJLST22}, as the model must be flexible enough in learning the new task while maintaining stability required to preserve existing knowledge.

Parameter-Efficient Fine-Tuning (PEFT) methods \citep{DBLP:conf/icml/HoulsbyGJMLGAG19} address the computational cost induced by full fine-tuning of LLMs for every new task. Based on the observation that model updates have a low intrinsic dimension \citep{DBLP:conf/iclr/LiFLY18, DBLP:conf/acl/AghajanyanGZ20}, Low-Rank Adaptation (LoRA) \citep{DBLP:journals/corr/abs-2106-09685} has emerged as a standard for CL alongisde its variants. LoRA approximates the weight update through the product of two trainable low-rank matrices, while freezing the original weight matrix, significantly reducing the memory footprint.

Despite their parameter efficiency, naively training low-rank adapters for each new task often leads to task interference, causing notable forgetting on earlier tasks \citep{DBLP:conf/cvpr/LiangL24, DBLP:conf/emnlp/WangCGXBZZGH23}. Current state-of-the-art CL methods address this by imposing constraints on the adapter updates. Subspace-partitioning methods restrict updates to lie orthogonal to previous tasks. For instance, O-LoRA \citep{DBLP:conf/emnlp/WangCGXBZZGH23} enforces orthogonality between successive low-rank matrices, while InfLoRA \citep{DBLP:conf/cvpr/LiangL24} projects gradients onto the orthogonal complement of previous task subspaces to eliminate interference. Alternatively, coordinate-wise methods such as ELLA \citep{DBLP:conf/eacl/BiswasZPBR26} restrict specific sets of parameters that can be modified when learning a new task. While methods like ELLA effectively mitigate forgetting, they operate in a dense parameter update regime, where all low-rank coordinates remain active for every task. A fundamental limitation of this dense approach is that it cannot achieve full parameter isolation. As every weight remains subject to optimization at each step, the update must navigate an increasingly constrained landscape to satisfy all previous alignment constraints simultaneously. As the task stream grows, this lack of structural separation can lead to capacity saturation or the accumulation of residual gradient interference, ultimately diminishing the model's ability to learn new information without forgetting the old.

\begin{figure}[t] %
    \centering
    \includegraphics[width=\linewidth]{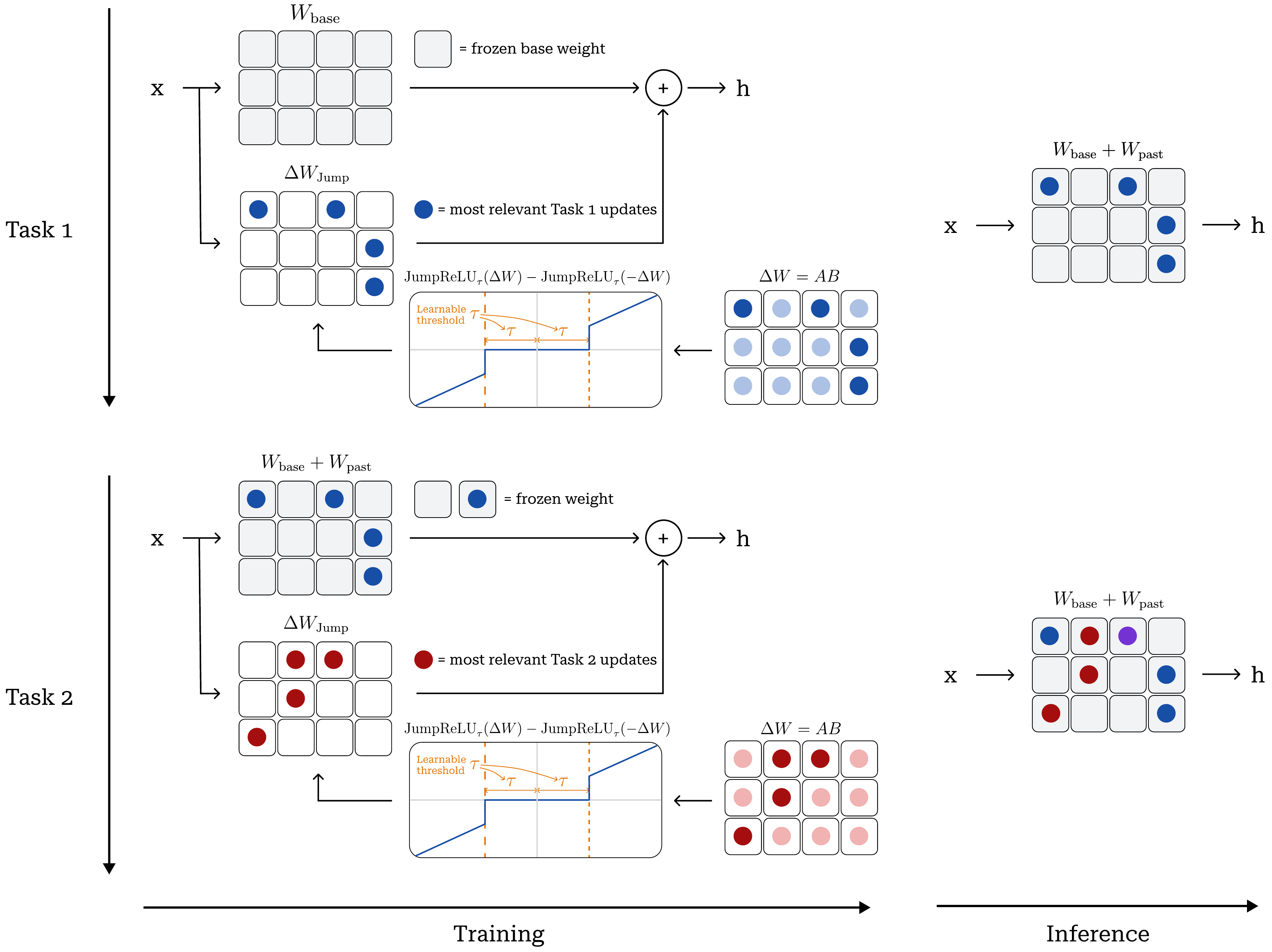}
    \caption{\textbf{Continual Learning with \jumplora{}.} We construct LoRA updates that are able to perform fine-grained interventions by repurposing the JumpReLU activation function such that it can be applied to weight updates during training. For each task we train a learnable threshold $\tau$ alongside the LoRA weights, meant to cut off low-magnitude updates, enabling adapters to specifically target only the most relevant parameters. This change effectively reduces the impact of the adapter upon the base weights while reducing the overlap between different task adapters.}
    \label{fig:main_architecture}
\end{figure}

In this paper, we introduce \jumplora, a method designed to achieve adaptive parameter isolation. Our approach leverages JumpReLU~\citep{DBLP:journals/corr/abs-2407-14435-jumpsae} as an activation function to adaptively induce sparsity on the LoRA blocks. Unlike traditional regularization methods, \jumplora \;dynamically cancels redundant or interfering weights, creating sparse adapters that minimize overlap with previous knowledge. We illustrate our approach in Figure \ref{fig:main_architecture}.

The primary contributions of our work are as follows:
\begin{itemize}
    \item We introduce \jumplora, which to the best of our knowledge, is the \emph{first framework to integrate learnable JumpReLU gating} for low-rank adapters. This mechanism enables precise, coordinate-wise sparsity in weight updates by optimizing a threshold directly alongside the adapter parameters.

    \item We adapt the \jumplora \;framework in a CL setup, which induces adaptive sparsity per task. This allows different adapters to occupy disjoint parameter coordinates and help mitigate task interference.
    
    \item We demonstrate the modularity and efficacy of \jumplora{} by applying it on top of existing CL frameworks. Through extensive benchmarking, we show that \jumplora{} improves upon existing approaches like IncLoRA and ELLA on the Standard CL Benchmark \citep{DBLP:conf/nips/ZhangZL15} and Long Sequence Benchmark \citep{DBLP:conf/iclr/RazdaibiedinaMH23}.
\end{itemize}

\section{Related Work}
Training neural networks on sequential tasks results in catastrophic forgetting \citep{mccloskey1989catastrophic}, where learning new tasks degrades the model's performance on previously learned tasks. Continual Learning (CL) is a research area that covers methods aiming to learn new tasks over time, while mitigating catastrophic forgetting.

CL approaches belong to 3 major categories: rehearsal-based, regularization-based and architecture-based. Rehearsal-based approaches \citep{DBLP:conf/nips/Lopez-PazR17,DBLP:journals/corr/abs-1906-01076,DBLP:conf/iclr/RiemerCALRTT19} store past examples in a replay buffer which is leveraged during training on the current task.

Regularization-based methods impose constraints on the optimization process to protect previous task knowledge. This is typically achieved with penalty terms or gradient constraints. A foundational approach is Elastic Weight Consolidation (EWC) \citep{DBLP:journals/corr/KirkpatrickPRVD16}, which computes the Fisher Information Matrix to establish parameter importance. Similarly, replay-free distillation is used to preserve prior tasks' outputs when dealing with new tasks~\citep{DBLP:journals/pami/LiH18a}. \cite{DBLP:conf/aistats/FarajtabarAML20} prevent interference by projecting the current updates onto a subspace orthogonal to the gradient of the previous tasks. Orthogonal isolation has then been extended to a PEFT CL setup, where O-LoRA \citep{DBLP:conf/emnlp/WangCGXBZZGH23} adds a loss term to penalize new LoRA adapters if they are not orthogonal to past adapters. InfLoRA \citep{DBLP:conf/cvpr/LiangL24} alternatively achives interference-free adaptation by projecting gradients onto the orthogonal complement of previous task subspaces, preventing new updates from interfering with earlier tasks without requiring an explicit orthogonality loss.

In addition to subspace-aware methods, other approaches perform coordinate-wise updates to achieve more granular isolation within shared parameters. ELLA~\citep{DBLP:conf/eacl/BiswasZPBR26} prevents interference with weight coordinates that had historically high-magnitude in previous tasks. Piggyback~\citep{DBLP:conf/eccv/MallyaDL18} builds on ideas from network quantization and pruning, adapting frozen backbone networks to new tasks by learning end-to-end binary masks that selectively activate existing weights, enabling continual learning without modifying shared parameters or suffering from catastrophic forgetting. Similarly, MIGU~\citep{DBLP:conf/emnlp/DuCLQHCC024} performs coordinate-wise isolation by masking gradients corresponding to weights connected to high-magnitude neurons, while HAT \citep{DBLP:conf/icml/SerraSMK18} learns task-specific, nearly binary attention masks via backpropagation to gate network units, conditioning gradients to freeze weights important for previous tasks, while dynamically allocating network capacity. Recently, \cite{zhang2025lori} prevent parameter interference by freezing the LoRA matrices $A$ as random projections and applying calibrated sparse masks $M$ to the expansion matrices $B$. The sparse masks $M$ are obtained by first fixing a sparsity ratio $s$. A task-specific threshold $\tau_t$ is then derived as the $s$-quantile of the parameter magnitudes in $B$ matrices, which are learned on an initial calibration set. Finally, the masks for $B$ are generated by keeping indices whose values are larger than the threshold $\tau_t$. In contrast, our approach, \jumplora, provides a more adaptive framework, by using a \emph{learnable JumpReLU gating} to obtain the sparse masks $M$. Thus, instead of relying on a fixed sparsity ratio $s$, our method changes the threshold $\tau_t$ to dynamically induce different levels of sparsity in the LoRA adapters, based on the task-specific complexity. 

Architecture-based methods prevent interference by isolating task knowledge within dedicated parameters or sub-networks. This can be achieved by dynamically reusing or expanding the underlying model's layers during training \citep{DBLP:conf/iclr/YoonYLH18,DBLP:conf/icml/LiZWSX19}, or by freezing previous task weights and progressively adding new columns with lateral connections for new tasks \citep{DBLP:journals/corr/RusuRDSKKPH16}. More recent approaches use Parameter-Efficient Fine-Tuning (PEFT) to avoid modifying the base model. This is achieved by learning task-specific soft prompts \citep{DBLP:conf/iclr/RazdaibiedinaMH23} or learning a new adapter for each task and select the relevant adapter at test time~\citep{DBLP:conf/acl/WangLJWWJCHWSZ23}. While these approaches isolate task-specific parameters, they require either knowing the Task ID at test time or defining a procedure to select the appropriate adapter for an unknown input. Our method merges existing task adapters into a single adapter, which allows tackling unseen tasks at test time without relying on Task IDs.

\section{Method}

\textbf{Continual Learning setup.} We consider the challenging rehearsal-free continual learning setting with task-agnostic inference for pretrained large language models. In this setting, the model has no access to data from previous tasks and must produce predictions without knowing the task identity of the input. Given a set of tasks  $\mathcal{T}$, for each supervised task $T \in \mathcal{T}$, defined as $T = \{(x^T_i, y^T_i)\}_{i=1}^{n_T}$, we train a low-rank adapter $\Delta W = A\cdot B$, with $A \in \mathbb{R}^{d_{\text{in}} \times r}$, $B \in \mathbb{R}^{r \times d_{\text{out}}}$ and $r \ll min(d_\text{in}, d_\text{out})$ for each base weight matrix $W_\text{base} \in \mathbb{R}^{d_{\text{in}} \times d_{\text{out}}}$. The output of the base layer $h = x \cdot W_\text{base}$ thus becomes $h = x \cdot (W_\text{base} + \Delta W)$.
Upon training on each task, the adapter is merged into the base weights: $W_\text{base} \gets W_\text{base} + \Delta W$.

\textbf{Proposed methodology.} Because $\Delta W$ is the product of two low-rank matrices, performing fine-grained interventions that target specific parameters of $W_\text{base}$ is by design difficult or even intractable. We therefore introduce a mechanism that enables the low-rank adapter to perform targeted updates on a task-relevant subset of parameters. Concretely, we operationalize task-relevance through gradient magnitude: parameters accumulating larger cumulative gradients are considered more important for the current task. Following \cite{DBLP:conf/emnlp/WangCGXBZZGH23}, we interpret $\Delta W$ as a proxy for the cumulative gradient of $W_\text{base}$ over the current task. We thus design a mechanism able to identify, retain and train only the tentative top-magnitude elements of $\Delta W$, effectively enabling sparse, targeted updates. We additionally impose the constraint that the sparsity level of each layer should not be treated as a tunable hyperparameter. Doing so would render the method prohibitively expensive, as it would require repeated training runs across all tasks to evaluate each candidate sparsity level. We design our method such that each layer can learn to select the level of sparsity that most appropriately delimits the relevant updates.

\textbf{JumpReLU.} We implement the proposed approach by repurposing the JumpReLU function, enabling its application on weight updates rather than activations. The JumpReLU function was introduced by \cite{DBLP:journals/corr/abs-2407-14435-jumpsae}, being formulated as follows:
\begin{equation}
    \text{JumpReLU}_{\tau}(x) = x\cdot H(x - \tau) ,\;\; \text{where} \;\; H(x)=\begin{cases}
        0 & \text{if } x \le 0 \\
        1 & \text{if } x > 0
        \end{cases}.
\end{equation}
Here, $\tau$ is a learnable threshold and $H$ is the Heaviside step function. Since the Heaviside function is discontinuous, its derivative is undefined at $x=\tau$ and zero almost everywhere else, making direct gradient-based optimization of $\tau$ infeasible. To address this, during backpropagation, a pseudo-derivative is computed using straight-through-estimators (STEs; \cite{DBLP:journals/corr/BengioLC13-STE}):
\begin{equation}
    \frac{\eth}{\eth \tau}\text{JumpReLU}_{\tau}(x)=
    -\frac{\tau}{\epsilon}\left(
        H\left(\frac{x-\tau}{\epsilon} + \frac{1}{2}\right) -
        H\left(\frac{x-\tau}{\epsilon} - \frac{1}{2}\right)
    \right).
\end{equation}
Within the context of this pseudo-derivative, the kernel bandwidth parameter $\epsilon$ controls the width of the region around $\tau$ within which inputs contribute to the gradient update of the threshold.

\textbf{{\jumplora}.} Applying JumpReLU to LoRA weight updates rather than pre-activations introduces several non-trivial challenges. First, the matrix $B$ is initialized to zero, meaning that $\Delta W = A\cdot B = \mathbf{0}$ at the start of training. Thresholding $\Delta W$ at any positive value $\tau > 0$ would therefore block all gradient flow at initialization, preventing the adapter from learning altogether. Note that a negative threshold is not a solution, since our objective is to retain the top-magnitude elements of $\Delta W$, regardless of their sign, which is equivalent to computing $\Delta W \odot H(|\Delta W| - \tau)$. Second, unlike pre-activations, which vary across inputs and thus stochastically fluctuate around the threshold, weight updates are deterministic given the current model state. Consequently, a parameter whose magnitude falls below $\tau$ at a given training step would stop receiving further gradient updates, recovering only if the underlying low-rank matrices shift enough to push them back above the threshold as a secondary effect. This issue is further compounded by the single-epoch-per-task training regime commonly used by current LoRA variants for CL (e.g., \citep{DBLP:conf/eacl/BiswasZPBR26}, \citep{DBLP:conf/emnlp/WangCGXBZZGH23}), which leaves insufficient time to reliably estimate an appropriate initial sparsity cut-off.

To address these challenges, we propose a gradual scheduling mechanism that defers and progressively introduces the JumpReLU sparsification. Concretely, rather than applying JumpReLU to $\Delta W$ from the outset, we define the effective weight update as a convex interpolation between full and sparse updates:
\begin{equation}
    \Delta W_\text{interp} = (1 - \gamma)\cdot \Delta W + \gamma\cdot(
        \text{JumpReLU}_\tau(\Delta W) -
        \text{JumpReLU}_\tau(-\Delta W)
    ),
\end{equation}
where $\gamma \in [0, 1]$ is a scheduling coefficient that is gradually annealed from $0$ to $1$ over the course of training. In the early stages ($\gamma \approx 0$), all parameters receive unrestricted gradient updates, allowing $\Delta W$ to develop meaningful structure before any sparsification is imposed. As $\gamma$ increases, the influence of the JumpReLU gate grows, progressively enforcing sparsity. Once $\gamma = 1$, the update reduces entirely to the sparsified form. This schedule ensures that parameters initially falling below the threshold are not prematurely discarded, retaining the ability to grow above $\tau$ if they prove to be task-relevant.

To further support stable threshold estimation, we introduce two scheduling hyperparameters, $S_{\text{start}}$ and $S_{\text{final}}$. During the first $S_{\text{start}}$ iterations, the adapter is trained without any sparsification, allowing $\Delta W$ to develop sufficient structure for a meaningful initial threshold to be computed. At step $S_{\text{start}}$, the threshold $\tau$ is initialized such that the number of elements in $|\Delta W|$ exceeding $\tau$ equals the total number of trainable parameters in $A$ and $B$. The interpolation procedure described above is then carried out over the interval $[S_{\text{start}}, S_{\text{final}}]$, with $\gamma$ annealed linearly from $0$ to $1$. After $S_{\text{final}}$ iterations, $\gamma = 1$ and the adapter is trained exclusively with the JumpReLU-sparsified update, fine-tuning only the selected top-magnitude parameters.

\begin{algorithm}[t]
\caption{Continual Learning with \jumplora}
\label{alg:jumplora}
\begin{algorithmic}[1]
\REQUIRE Tasks $\mathcal{T} = \{T_1, \dots, T_n\}$, base weights $W_\text{base}$, LoRA rank $r$, LoRA scale $\alpha$, bandwidth $\epsilon$, first interpolation step $S_\text{start}$, final interpolation step $S_\text{final}$
\STATE (Optional) Initialize ELLA state: $W_\text{past} \gets 0$
\FOR{each task $T_i \in \mathcal{T}$}
    \STATE Inject \jumplora{} layers: $\Delta W = A\cdot B$, where $A \in \mathbb{R}^{d_\text{in} \times r}$ and $B \in \mathbb{R}^{r \times d_\text{out}}$
    \STATE Initialize $A \sim \text{KaimingUniform}$, $B \gets \mathbf{0}$ and the LoRA scaling factor $\beta \gets \alpha/r$ 
    \STATE Set interpolation factor $\gamma \gets 0$ 
    
    \FOR{each training step $s$}
        \IF{$s = S_\text{start}$}
            \STATE Initialize threshold $\tau$ such that the number of parameters from $|\Delta W|$ that are above the threshold is equal to the number of total parameters from $A$ and $B$.
        \ENDIF
        
        \STATE Update $\gamma \gets \min\left(1.0, \max\left(0.0, \frac{s - S_\text{start}}{S_\text{final} - S_\text{start}}\right)\right)$ 
        
        \STATE Compute dense update: $\Delta W \gets A\cdot B$ 
        \STATE Compute sparse update: $\Delta W_\text{jump} \gets \text{JumpReLU}_\tau(\Delta W) -
        \text{JumpReLU}_\tau(-\Delta W)$ 
        \STATE Interpolate: $\Delta W_\text{interp} \gets \gamma \cdot \Delta W_\text{jump} + (1 - \gamma) \cdot \Delta W$ 
        
        \STATE Forward Pass: $h \gets x\cdot (W_\text{base} + \beta\cdot \Delta W_\text{interp})$ 
        \STATE Compute Loss $\mathcal{L}$ 
        \STATE (Optional) ELLA penalty: $\mathcal{L} \gets \mathcal{L} + \lambda \|\Delta W \odot W_\text{past}\|_F^2$ if $s < S_\text{start}$ else $\mathcal{L} \gets \mathcal{L} + \lambda \|\Delta W_\text{jump} \odot W_\text{past}\|_F^2$
        \STATE Update $\{A, B, \tau\}$ via backpropagation
    \ENDFOR
    
    \STATE Compute final sparse adapter: $\Delta W_\text{final} \gets (A \cdot B) \odot H(|A\cdot B| - \tau)$ 
    \STATE Merge adapter: $W_\text{base} \leftarrow W_\text{base} + \beta \cdot \Delta W_\text{final}$ 
    \STATE (Optional) Update ELLA state: $W_\text{past} \leftarrow W_\text{past} + \Delta W_\text{final}$ 
    \STATE Remove task-specific adapters $\Delta W$ from the model
\ENDFOR
\end{algorithmic}
\label{jumplora_algo}
\end{algorithm}

\textbf{Implementation.} Algorithm~\ref{alg:jumplora} provides a complete description of the \jumplora \;continual learning procedure. At the beginning of each task $T_i$ (lines 3-5), a fresh LoRA adapter $\Delta W = A\cdot B$ is injected into the model, with $A$ initialized via Kaiming uniform initialization, $B$ initialized to zero, and the interpolation factor set to $\gamma = 0$. The LoRA scaling factor is fixed to $\lambda = \alpha / r$ throughout training.

At each training step $s$, $\gamma$ is updated at line 10 according to:
\begin{equation}
    \gamma = \text{clip}\left(\frac{s - S_{\text{start}}}{S_{\text{final}} - S_{\text{start}}}, 0, 1\right),
\end{equation}
ensuring that $\gamma$ remains in $[0, 1]$ and increases linearly over the interpolation interval. At step $S_{\text{start}}$, the threshold $\tau$ is initialized as described above (lines 7-9). The effective weight update is then computed in three stages (lines 11-13). First, the dense update $\Delta W = A\cdot B$ is computed. Second, a sparse update is obtained by applying JumpReLU symmetrically to account for both positive and negative entries of $\Delta W$:
\begin{equation}
    \Delta W_{\text{jump}} = \text{JumpReLU}_\tau(\Delta W) - \text{JumpReLU}_\tau(-\Delta W).
\end{equation}
Third, the two updates are interpolated as $\Delta W_{\text{interp}} = \gamma \cdot \Delta W_{\text{jump}} + (1 - \gamma) \cdot \Delta W$. The forward pass is then computed as $h = x\cdot (W_{\text{base}} + \lambda \cdot \Delta W_{\text{interp}})$ (line 14), and the model is optimized with respect to the task loss, with $A$, $B$, and $\tau$ all updated via backpropagation (lines 15-17). Within this context, we also showcase the fact that \jumplora{} can easily be combined with other LoRA-based continual learning procedures. For instance, at line 16, we optionally add the ELLA \citep{DBLP:conf/eacl/BiswasZPBR26} regularization penalty.

Upon completion of training on task $T_i$, the final sparse adapter is obtained by hard-thresholding at line 19: $\Delta W_{\text{final}} = (A\cdot B) \odot H(|A\cdot B| - \tau)$, and merged into the base weights at line 20 as $W_{\text{base}} \gets W_{\text{base}} + \lambda \cdot \Delta W_{\text{final}}$. The task-specific adapter is then discarded (line 22). Optionally, the ELLA state is updated at line 21 by accumulating $\Delta W_{\text{final}}$ into $W_{\text{past}}$, providing a record of parameters modified in previous tasks to inform future regularization. Throughout this procedure, we employ the JumpReLU implementation provided by \cite{DBLP:journals/corr/abs-2407-14435-jumpsae}.

\section{Experiments}

\subsection{Datasets}
We evaluate our models using the Standard CL Benchmark~\citep{DBLP:conf/nips/ZhangZL15}, which consists of five text classification datasets corresponding to two tasks: sentiment analysis (Yelp and Amazon Reviews) and topic classification (DBPedia, Yahoo and AGNews). Following \cite{DBLP:conf/emnlp/WangCGXBZZGH23}, we use four datasets (AGNews, Amazon, DBpedia, and Yahoo) and adopt the three task sequence orders that they used in their study (named Order 1, 2 and 3).

We also evaluate our models on the Long Sequence Benchmark~\citep{DBLP:conf/iclr/RazdaibiedinaMH23}, which consists of 15 datasets: the original five datasets from Standard CL, four datasets from the GLUE benchmark~\citep{DBLP:conf/iclr/WangSMHLB19} (MNLI, QQP, RTE, SST2), five datasets from the SuperGLUE benchmark~\citep{DBLP:conf/nips/WangPNSMHLB19}  (WiC, CB, COPA, MultiRC, BoolQ), and the IMDB movie reviews dataset~\citep{DBLP:conf/acl/MaasDPHNP11}. We also adopt the dataset orders in their sequential training, named Order 4, 5 and 6. Following \cite{DBLP:conf/eacl/BiswasZPBR26}, we keep a maximum of 500 samples per class for the test set. We collect a combined training and validation pool, by sampling at most 1000 examples per class, which are then partitioned into an 80\%/20\% train-validation split. The detailed datasets and orders are shown in Appendix~\ref{app:datasets}.

\subsection{Metrics}
Let $|\mathcal{T}|$ be the number of tasks, $a_{0,i}$ be the performance of task i when trained in isolation and $a_{i,j}$ be the performance of task j after training on task i. We evaluate the following metrics: Overall Accuracy (OA)~\citep{DBLP:conf/eccv/ChaudhryDAT18}, Backward Transfer (BWT)~\citep{DBLP:journals/corr/abs-2211-12701} and Forward Transfer (FWT)~\citep{DBLP:conf/nips/Lopez-PazR17}. OA measures the average performance on each task after finishing training on the final task $|\mathcal{T}|$: $\text{OA}=\frac{1}{|\mathcal{T}|}\sum_{i=1}^{|\mathcal{T}|}a_{|\mathcal{T}|,i}$. BWT measures how much the performance of subsequent tasks affect the performance of prior tasks: 
$\text{BWT}=\frac{1}{|\mathcal{T}|-1}\sum_{i=1}^{|\mathcal{T}|-1}(a_{|\mathcal{T}|,i}-a_{i,i})$. Finally, FWT measures how much previous tasks improve a new task: $\text{FWT}=\frac{1}{|\mathcal{T}|}\sum_{i=1}^{|\mathcal{T}|}(a_{i,i}-a_{0,i})$.

\subsection{Baselines}
In all our experiments, we apply PEFT-based continual learning strategies on top of the T5 model~\citep{DBLP:journals/jmlr/RaffelSRLNMZLL20}, which uses an encoder-decoder architecture with 770M parameters. We employ encoder-decoder models as they have shown to outperform similarly-sized decoder-only models in the context of CL benchmarks~\citep{DBLP:conf/eacl/BiswasZPBR26}.

Our \jumplora{} framework is modular and can be integrated with existing PEFT-based CL approaches. To evaluate its efficacy, we combine it with two primary baselines: i) IncLoRA, a high-plasticity baseline where new adapters are learned sequentially on each incoming task, and no replay or regularization is used; ii) ELLA, the current state-of-the-art method on the two benchmarks. We report numbers for the both standalone baselines as well as \jumplora-enhanced versions (\jumplora{}+IncLoRA and \jumplora{}+ELLA). Since code repositories are not officially released, all reported numbers are from our own implementations, with the exception of the classical EWC method, which is included as reported by \cite{DBLP:conf/eacl/BiswasZPBR26}.

\subsection{Implementation details}
All experiments were conducted on compute nodes with 8 Nvidia H200 GPUs. Our implementation uses the Huggingface ecosystem, specifically the Transformers v4.57.6~\citep{DBLP:conf/emnlp/WolfDSCDMCRLFDS20} and PEFT v0.18.1~\citep{peft} libraries. We report the mean performance across three independent runs, each initialized with a unique random seed (42, 43 and 44). 

For all orders in both the SC and LS Benchmarks, we train each task with one epoch, using AdamW~\citep{DBLP:conf/iclr/LoshchilovH19} and the WarmupLR scheduler with a total batch size of $32$. We used a constant learning rate of $10^{-3}$. 
Following \cite{DBLP:journals/corr/abs-2407-14435-jumpsae}, we set $\epsilon = 0.001$. Additionally, we set $S_\text{start}$ to $20\%$ of an epoch and $S_\text{final}$ to $80\%$ of an epoch for all experiments. Following \cite{DBLP:conf/acl/WangLJWWJCHWSZ23}, all LoRA adapters are applied only on the query and value matrices of each attention block, while the adapter rank is set to $r=8$ and the scale to $\alpha=32$. The ELLA-based models require tuning the parameter $\lambda$ that controls the penalty term. In Appendix~\ref{app:hyperparameters}, we list the $\lambda$ values for all orders and experiments. We discuss several design choices for our method in Section \ref{ablation_section}. The reported results in Section \ref{main_results_section} correspond to the best configuration for each benchmark.

\subsection{Results} \label{main_results_section}
\begin{table*}[!ht]
\caption{Overall Average Accuracy (OA) comparison of the first four baselines on Standard CL benchmark (Order $1, 2, 3$) and Long Sequence benchmark (Order $4, 5, 6$). Results are averaged across three seeds. Top scores are in bold.}
\vspace{-6pt}
\centering
{
\begin{tabular}{lcccccccc}
\toprule
\multirow{2}{*}{\textbf{Methods}} & \multicolumn{4}{c}{\textbf{\textcolor{datablue}{{Standard CL Benchmark (SC)}}}} & \multicolumn{4}{c}{\textbf{\textcolor{datablue}{Long Sequence Benchmark (LS)}}} \\
& Order 1 & Order 2 & Order 3 & \textcolor{datablue}{OA} & Order 4 & Order 5 & Order 6 & \textcolor{datablue}{OA} \\
\midrule
EWC    & 46.30 & 45.30 & 52.10 & 47.90    & 44.90 & 44.00 & 45.40 & 44.80 \\
IncLoRA   & 59.25 & 58.69 & 69.86 & 62.60 & 56.83 & 56.40 & 54.32 & 55.89 \\
\jumplora{} + IncLoRA & 70.76 & 69.26 & 74.78 & 71.60 & 65.35 & 61.84 & 64.06 & 63.75  \\
ELLA & 78.09 & 78.37 & 78.23 & 78.23 & 72.64 & 67.91 & 74.15 & 71.57  \\
\jumplora{} + ELLA  & \textbf{78.84} & \textbf{79.05} & \textbf{78.67} & \textbf{78.85} & \textbf{73.60} & \textbf{68.76} & \textbf{75.81} & \textbf{72.72}  \\
\bottomrule
\end{tabular}
}
\label{tab:main_results}
\end{table*}


\begin{table*}[!ht]
\caption{Backward and forward transfer (BWT and FWT) on average for Standard CL benchmark (mean of order $1, 2, 3$) and Long Sequence benchmark (mean of order $4, 5, 6$). Results are averaged across three seeds. Top scores are in bold.}
\vspace{-6pt}
\centering
{
\begin{tabular}{lcccc}
\toprule
\multirow{2}{*}{\textbf{Methods}} & \multicolumn{2}{l}{\textbf{\textcolor{datablue}{{Standard CL Benchmark (SC)}}}} & \multicolumn{2}{l}{\textbf{\textcolor{datablue}{Long Sequence Benchmark (LS)}}} \\
& BWT & FWT & BWT & FWT \\
\midrule

IncLoRA   & -22.9 & -17.1 & -22.1 & -21.2  \\
{\jumplora} + IncLoRA & -11.9 & -8.1 & -15.5 & -13.3 \\
ELLA  & \textbf{-0.5} & -1.5 & -4.8 & -5.5 \\
{\jumplora} + ELLA & -1.9 & \textbf{-0.8} & \textbf{-4.5} & \textbf{-4.4} \\
\bottomrule
\end{tabular}
}
\vspace{-4pt}
\label{tab:fwt_bwt}
\end{table*}

%

\begin{figure}[t] %
    \centering
    \includegraphics[width=\linewidth]{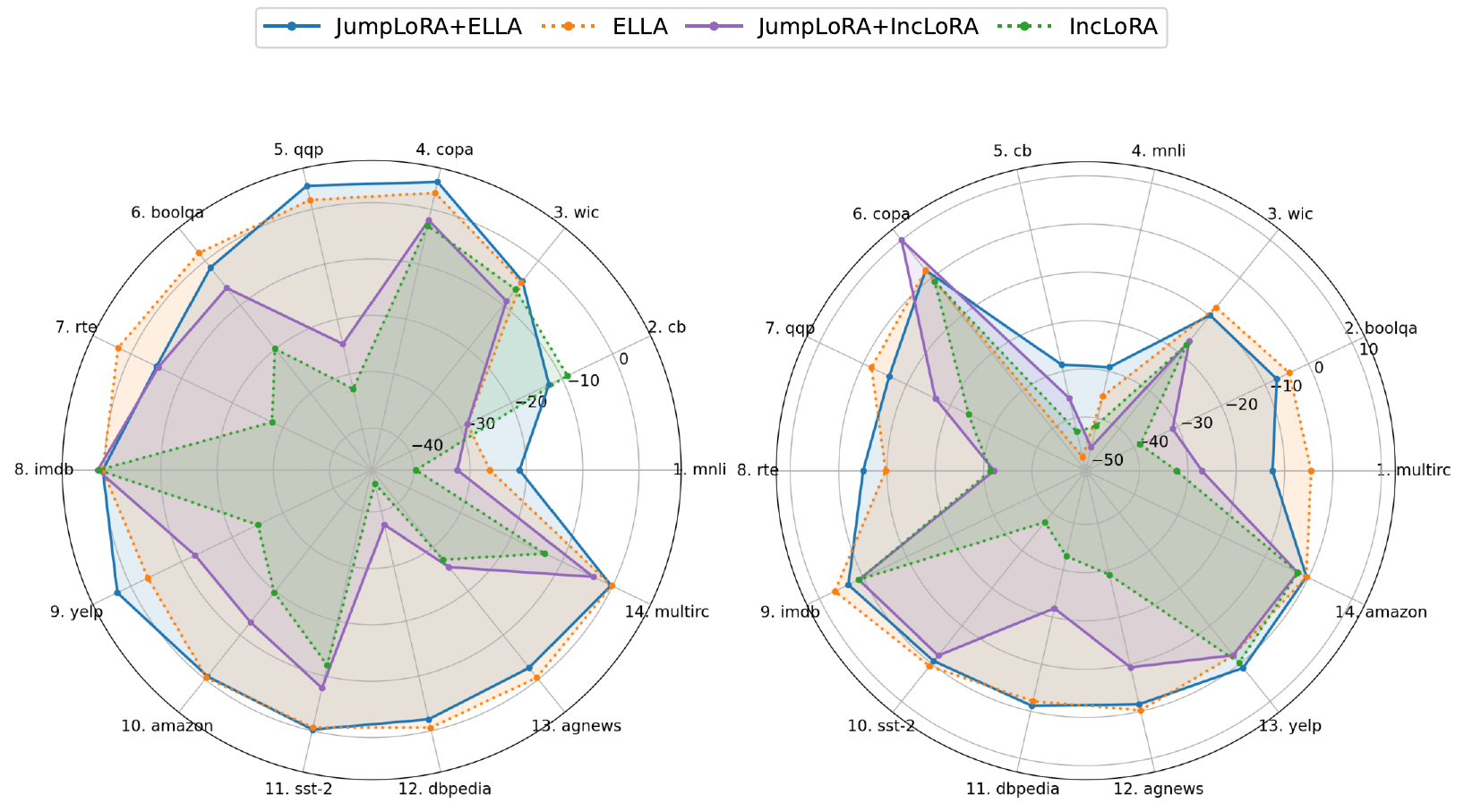}
    \caption{BWT scores during training for Orders 4 and 5. Comparison between base ELLA, IncLoRA and our \jumplora{} variants. Higher values indicate indicate less forgetting, i.e., better retention of previously learned tasks. Best viewed in color.}
    \label{fig:bwt_radar}
\end{figure}

We report overall accuracy in Table \ref{tab:main_results}, and forward and backward transfer in Table \ref{tab:fwt_bwt}, respectively. In terms of OA, {\jumplora} brings substantial improvements over the IncLoRA baseline. Notably, these gains are achieved with no explicit regularization to prevent task interference while training sequentially, indicating that the sparsity induced by {\jumplora} is sufficient for effective parameter isolation. Furthermore, {\jumplora} induces less forgetting and shows better knowledge reuse across tasks, as reflected by improved BWT and FWT scores, respectively. This behavior is further illustrated in Figure \ref{fig:bwt_radar}, which presents radial BWT profiles for two representative task orders. While the gains for {\jumplora} combined with ELLA are more modest, the improvements for {\jumplora}+IncLoRA are consistently more pronounced, showing a clearer reduction in forgetting across tasks. Notably, certain sharp drops (e.g., for MNLI and CB) appear across all methods in these orders, suggesting that they may be driven by task-specific difficulty rather than catastrophic forgetting. Overall, the improvements remain consistent across both short and long CL benchmarks, highlighting the robustness of our method.

Combined with ELLA, which explicitly controls task interference, {\jumplora} consistently yields better or competitive performance across both SC and LS benchmarks compared to standalone ELLA, further advancing the state-of-the-art. In particular, {\jumplora}+ELLA achieves the highest OA on all six orders, as well as the best FWT and BWT scores on the LS benchmark. On shorter task sequences, its performance approaches that of training each task independently, as evidenced by the near-zero FWT scores. Overall, the results indicate that the adaptive sparsity induced by {\jumplora} is complementary to, and can effectively enhance, regularization-based approaches.

\subsection{Ablations} \label{ablation_section}
We perform several ablations on our JumpReLU framework. First, we evaluate two settings for the threshold: \textbf{i) global:} a single sparsity threshold $\tau$ for all the layers, and \textbf{ii) local:} there is a different sparsity threshold $\tau$ per transformer block. Employing a per-block learnable threshold should in theory provide the model with greater flexibility in adjusting the sparsity according to the contribution of transformer blocks at different model depths.

Secondly, for \jumplora{} applied to ELLA, we evaluate two settings: \emph{sparse update} (the implementation in Algorithm~\ref{jumplora_algo}) and \emph{interpolated update}, where the element-wise ELLA penalty loss is computed w.r.t.~the interpolated update $\Delta W_{\text{interp}}$, instead of the full sparse update $\Delta W_{\text{jump}}$.

We report the mean OA across orders for the SC and LS benchmarks in Table \ref{tab:ablations}, for all ablation configurations. In general, the results demonstrate that {\jumplora} is robust to different design choices. For {\jumplora}+IncLoRA, using a per-block (local) sparsity threshold improves performance on the SC benchmark, although this trend does not transfer to the LS benchmark, where the global threshold recovers performance. Across configurations, sparse updates consistently provide an advantage over interpolated ones, though both remain effective. When combined with ELLA, {\jumplora} maintains stable performance under all settings, with the best results achieved using a global threshold with sparse updates on SC, and a local threshold with sparse updates on LS.



\begin{table}[t]
\caption{Ablation study regarding the combinations of global/local threshold the sparse/interpolated updates and their overall accuracy (OA) for Standard CL benchmark (Order $1, 2, 3$) and Long Sequence benchmark (Order $4, 5, 6$).}
\vspace{-6pt}
\centering
\label{tab:summary_threshold_update}
\setlength{\tabcolsep}{6pt}
\begin{tabular}{llccc}
\toprule
Method & Threshold & Update & \makecell{\textbf{\textcolor{datablue}{{Standard CL Benchmark}}} \\ OA} 
& \makecell{\textbf{\textcolor{datablue}{Long Sequence Benchmark}} \\ OA} \\
\midrule

\multirow{2}{*}{{\jumplora} + IncLoRA}
  & Global & --     & 68.83          & 63.75          \\
  & Local  & --     & 71.60          & 62.96          \\

\midrule

\multirow{4.5}{*}{{\jumplora} + ELLA}
  & \multirow{2}{*}{Global} & Sparse & \textbf{78.85} & 71.90          \\
  &                  & Interp. & 78.54          & 72.15          \\
  \cmidrule(l){2-5}
  & \multirow{2}{*}{Local}  & Sparse & 78.48          & \textbf{72.72} \\
  &                  & Interp. & 78.46          & 72.31             \\

\bottomrule
\end{tabular}

\label{tab:ablations}
\end{table}

\section{Sparsity Analysis}

\begin{figure}[t] %
    \centering
    \includegraphics[width=\linewidth]{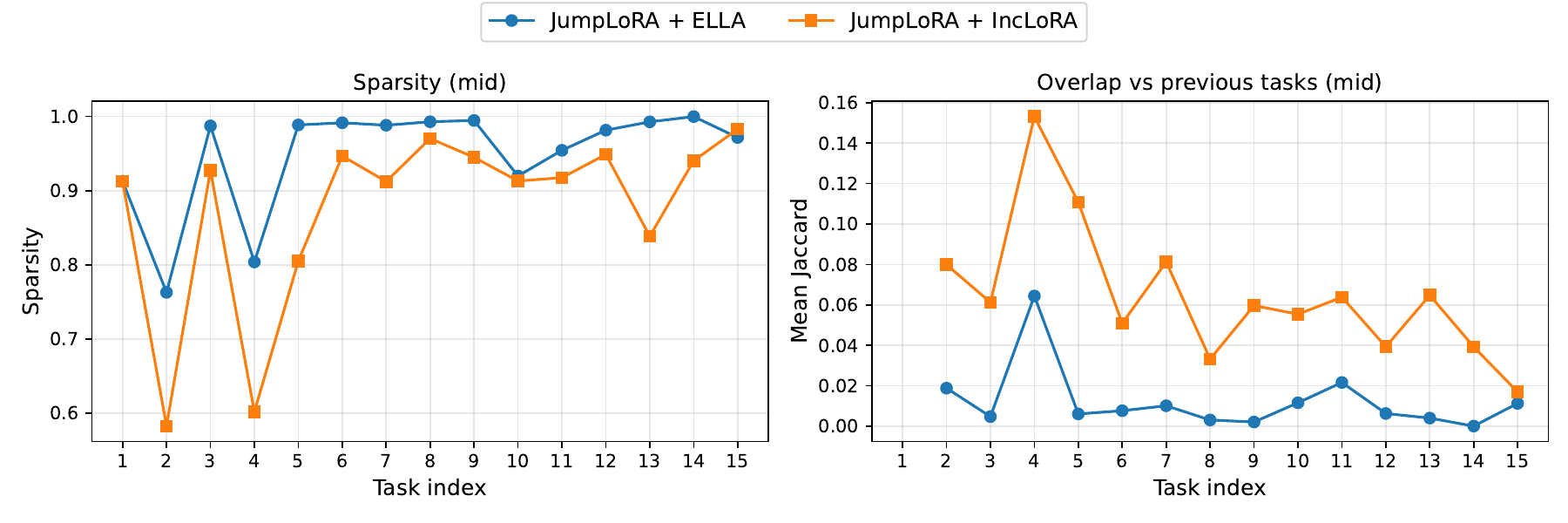}
    \caption{Sparsity and Average Jaccard overlap with the previous tasks on Long Order 4 for the middle layer. Best viewed in color.}
    \label{fig:sparsity_and_overlap}
\end{figure}

\begin{figure}[t] %
    \centering
    \includegraphics[width=\linewidth]{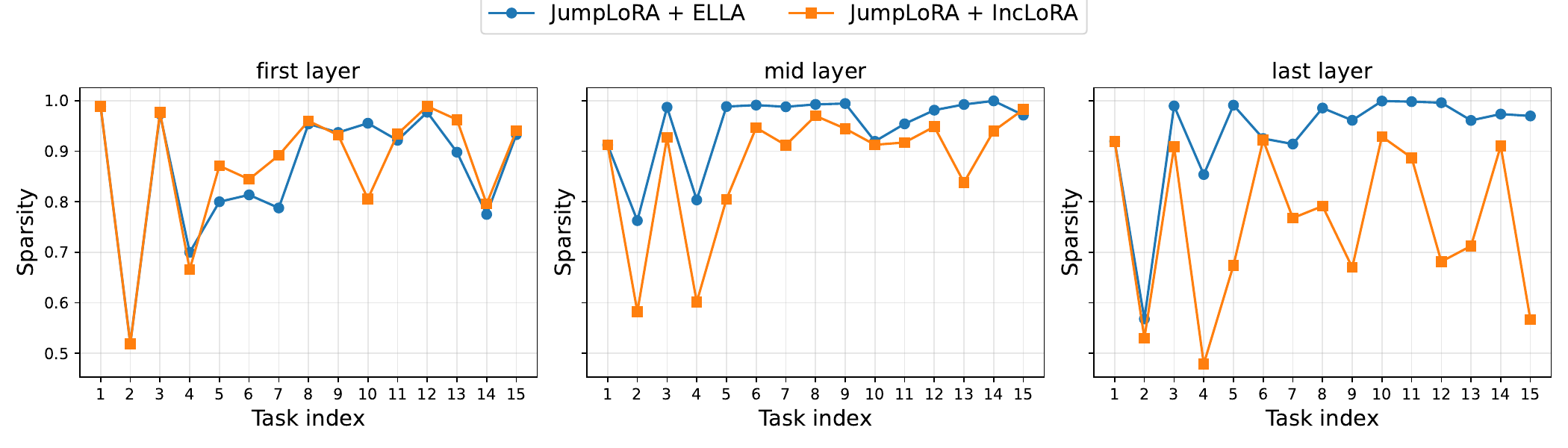}
    \caption{\textbf{Sparsity} comparison for IncLoRA and ELLA on Long Order 4 for the middle layer. Best viewed in color.}
    \label{fig:sparsity_across_layers}
\end{figure}

In this section, we evaluate the JumpReLU approach by analyzing two key metrics: sparsity and adapter overlap. Here, sparsity represents the fraction of elements in $\Delta W_\text{jump}$ zeroed by the learned threshold. We measure overlap via Jaccard similarity between binary supports, as a proxy for parameter interference versus reuse of the same parameters. Results for the middle layer of the model are illustrated in Figure \ref{fig:sparsity_and_overlap}. We notice that \jumplora\;induces high sparsity in both models. The additional coordinate-wise loss penalty of ELLA results in a higher 94.9\% average sparsity, as compared to the 87.6\% average sparsity for IncLoRA. Moreover, the average Jaccard overlap is small (0.012 for ELLA and 0.065 for IncLoRA). Together, these results indicate that \jumplora{} induces a high degree of parameter isolation and helps identify disjoint task-specific weight updates, which can be merged with negligible interference.

In Figure \ref{fig:sparsity_across_layers}, we additionally compare the evolution of sparsity for \jumplora{} applied on top of IncLoRA and ELLA with model depth.
We observe that the ELLA loss term helps maintain a higher degree of sparsity for the deeper layers, consistent with the improved performance observed for \jumplora{}+ELLA.

\section{Conclusions}
In this paper, we introduced \jumplora{}, a novel framework that repurposes JumpReLU to optimize a learnable threshold and achieve adaptive sparsity in LoRA weights. We demonstrate the effectiveness and modularity of our framework by integrating it with PEFT-based CL methods. Our empirical results indicate that \jumplora{} brings significant performance boosts for IncLoRA, and improves the state-of-the-art performance of ELLA. We further show that our method results in high overall sparsity and low overlap between adapter weights, therefore effectively targeting task interference via parameter isolation. Overall, \jumplora{} presents itself as a flexible approach for PEFT-based continual learning, bringing performance gains by enhancing task decoupling via learnable sparsity.

While in this study we focused on the CL setup for natural language processing tasks, our framework is flexible and can be applied to other domains powered by PEFT approaches. Future work will tackle the application of \jumplora{} in a CL setup for computer vision and evaluate the efficiency of our learnable threshold mechanism for sparse low-rank updates in vision transformers~\citep{DBLP:conf/iclr/DosovitskiyB0WZ21}. 

\bibliography{collas2026_conference}
\bibliographystyle{collas2026_conference}

\clearpage

\appendix
\section{Appendix}
\subsection{Datasets and Task Orders}
\label{app:datasets}

\begin{table*}[!t]
\centering
\caption{
The five classification datasets in the Standard CL Benchmark~\citep{DBLP:conf/nips/ZhangZL15}.}
\begin{tabular}{lllll}
\toprule
\textbf{Dataset Name} & \textbf{Category} & \textbf{Task} & \textbf{Domain} \\
\midrule
Yelp       & CL Benchmark & Sentiment Analysis               & Yelp Reviews             \\
Amazon     & CL Benchmark & Sentiment Analysis               & Amazon Reviews           \\
DBPedia    & CL Benchmark & Topic Classification             & Wikipedia                \\
Yahoo      & CL Benchmark & Topic Classification             & Yahoo Q\&A               \\
AG News    & CL Benchmark & Topic Classification             & News                     \\
\bottomrule
\end{tabular}
\label{tab:shortseq_datasets}
\end{table*}

\begin{table*}
\centering
\caption{
The 15 classification datasets in the Long Sequence Benchmark~\citep{DBLP:conf/iclr/RazdaibiedinaMH23}.} 
\begin{tabular}{lllll}
\toprule
\textbf{Dataset Name} & \textbf{Category} & \textbf{Task} & \textbf{Domain} \\
\midrule
Yelp       & CL Benchmark & Sentiment Analysis               & Yelp Reviews             \\
Amazon     & CL Benchmark & Sentiment Analysis               & Amazon Reviews           \\
DBPedia    & CL Benchmark & Topic Classification             & Wikipedia                \\
Yahoo      & CL Benchmark & Topic Classification             & Yahoo Q\&A               \\
AG News    & CL Benchmark & Topic Classification             & News                     \\
MNLI       & GLUE         & Natural Language Inference       & Various                  \\
QQP        & GLUE         & Paragraph Detection              & Quora                    \\
RTE        & GLUE         & Natural Language Inference       & News, Wikipedia          \\
SST-2      & GLUE         & Sentiment Analysis               & Movie Reviews            \\
WiC        & SuperGLUE    & Word Sense Disambiguation        & Lexical Databases        \\
CB         & SuperGLUE    & Natural Language Inference       & Various                  \\
COPA       & SuperGLUE    & Question and Answering           & Blogs, Encyclopedia      \\
BoolQA     & SuperGLUE    & Boolean Question and Answering   & Wikipedia                \\
MultiRC    & SuperGLUE    & Question and Answering           & Various                  \\
IMDB       & SuperGLUE    & Sentiment Analysis               & Movie Reviews            \\
\bottomrule
\end{tabular}
\label{tab:longseq_datasets}
\end{table*}

\begin{table*}[t]
\centering
\caption{
Task sequence orders for both SC and LS Benchmarks. 
}
\begin{tabular}{lcp{10cm}}
\toprule
\textbf{Benchmark} & \textbf{Order} & \textbf{Task Sequence} \\
\midrule
\multirow{4}{*}{Standard CL Benchmark} 
  & 1 & dbpedia $\rightarrow$ amazon $\rightarrow$ yahoo $\rightarrow$ ag \\
  \cmidrule{2-3}
  & 2 & dbpedia $\rightarrow$ amazon $\rightarrow$ ag $\rightarrow$ yahoo \\
  \cmidrule{2-3}
  & 3 & yahoo $\rightarrow$ amazon $\rightarrow$ ag $\rightarrow$ dbpedia \\
\midrule
\multirow{7}{*}{Long Sequence Benchmark} 
  & 4 & mnli $\rightarrow$ cb $\rightarrow$ wic $\rightarrow$ copa $\rightarrow$ qqp $\rightarrow$ boolqa $\rightarrow$ rte $\rightarrow$ imdb $\rightarrow$ yelp $\rightarrow$ amazon $\rightarrow$ sst-2 $\rightarrow$ dbpedia $\rightarrow$ ag $\rightarrow$ multirc $\rightarrow$ yahoo \\
  \cmidrule{2-3}
  & 5  & multirc $\rightarrow$ boolqa $\rightarrow$ wic $\rightarrow$ mnli $\rightarrow$ cb $\rightarrow$ copa $\rightarrow$ qqp $\rightarrow$ rte $\rightarrow$ imdb $\rightarrow$ sst-2 $\rightarrow$ dbpedia $\rightarrow$ ag $\rightarrow$ yelp $\rightarrow$ amazon $\rightarrow$ yahoo \\
  \cmidrule{2-3}
  & 6 & yelp $\rightarrow$ amazon $\rightarrow$ mnli $\rightarrow$ cb $\rightarrow$ copa $\rightarrow$ qqp $\rightarrow$ rte $\rightarrow$ imdb $\rightarrow$ sst-2 $\rightarrow$ dbpedia $\rightarrow$ ag $\rightarrow$ yahoo $\rightarrow$ multirc $\rightarrow$ boolqa $\rightarrow$ wic \\
\bottomrule
\end{tabular}
\label{tab:task_orders}
\end{table*}

In Table~\ref{tab:shortseq_datasets}, we list the original five datasets in the Standard CL benchmark (SC). Following \cite{DBLP:conf/acl/WangLJWWJCHWSZ23}, we only used the last four datasets.

In Table~\ref{tab:longseq_datasets}, we list the 15 datasets in the Long Sequence Benchmark (LS). 

We report all task orders used for our CL experiments
in Table~\ref{tab:task_orders}, which are adopted in previous works~\citep{DBLP:conf/eacl/BiswasZPBR26}. Orders 1-3 refer to the Standard CL Benchmark, while orders 4-6 refer to the Long Sequence Benchmark.

\subsection{Hyperparameters}
Regarding $\lambda$, we consider two strategies: using the default values from the ELLA paper \citep{DBLP:conf/eacl/BiswasZPBR26} (in which the authors report that their $\lambda$ values were chosen to be the ones that balance the cross-entropy vs.~ELLA loss to a ratio closer to $1$), and performing a custom grid search in which, at each task, we maximize validation accuracy over both past and current tasks. The search range is set to $[1, 10, 10^2, 10^3, 10^4, 10^5]$. Although the best results reported in Table \ref{tab:main_results} are achieved with the default values, the ablation study in Table \ref{tab:ablations} shows that, under the global threshold setting and for longer task sequences, the best performance is obtained with the $\lambda$ values selected via grid search. Table \ref{tab:lambda_settings} reports the $\lambda$ values used for Table \ref{tab:main_results}, while Table \ref{tab:lambda_settings_ablations} lists those used for Table \ref{tab:ablations}, along with whether they correspond to the default or grid-searched setting.

The values obtained via grid search are often significantly lower than the default ones, suggesting that, for our method, less regularization from the ELLA loss is required. This may indicate that \jumplora already provides sufficient stability for certain tasks, reducing the need for stronger ELLA-based regularization.

\label{app:hyperparameters}
\begin{table*}[ht]
\centering
\caption{Hyperparameter settings for the main experiments in Table \ref{tab:main_results}.}
{%
\begin{tabular}{llll}
\toprule
\textbf{Experiment} & \textbf{Param Type} & \textbf{Order} & \textbf{$\lambda$ values} \\
\midrule
ELLA & default & 1--3 & $0, 3\times10^{4},\,\ldots,\,3\times10^{4}$ \\
\midrule
ELLA & default & 4 & $0,\,5{\times}10^{5},\,\ldots,\,5{\times}10^{5},\,5{\times}10^{7}$ \\
\midrule  
ELLA & default & 5 & $0,\,5{\times}10^{6},\,\ldots,\,5{\times}10^{6},\,5{\times}10^{7},\,5{\times}10^{7},\,5{\times}10^{7}$ \\
\midrule
ELLA & default & 6 & $0,\,5{\times}10^{5},\,\ldots,\,5{\times}10^{5}$ \\
\midrule
\jumplora\;+ ELLA & default & 1--3 & $0, 3\times10^{4},\,\ldots,\,3\times10^{4}$ \\
\midrule
\jumplora\;+ ELLA & default & 4 & $0,\,5{\times}10^{5},\,\ldots,\,5{\times}10^{5},\,5{\times}10^{7}$ \\
\midrule
\jumplora\;+ ELLA & default & 5 & $0,\,5{\times}10^{6},\,\ldots,\,5{\times}10^{6},\,5{\times}10^{7},\,5{\times}10^{7},\,5{\times}10^{7}$ \\
\midrule
\jumplora\;+ ELLA & default & 6 & $0,\,5{\times}10^{5},\,\ldots,\,5{\times}10^{5}$ \\

\bottomrule
\end{tabular}}
\label{tab:lambda_settings}
\end{table*}


\begin{table*}[ht]
\caption{Hyperparameter settings for the \jumplora+ELLA ablation experiments in Table \ref{tab:ablations}.}
\centering
\resizebox{0.98\textwidth}{!}{%
\begin{tabular}{lllll}
\toprule
\textbf{Threshold} & \textbf{Update} & \textbf{Param Type} & \textbf{Order} & \textbf{$\lambda$ values} \\
\midrule
 global & sparse & default & 1--3 & $0, 3\times10^{4},\,\ldots,\,3\times10^{4}$ \\
\midrule
 global & sparse & grid & 4 & $0, 10^3, 10^5, 10^5, 10^5, 10^2, 10^4, 10^5, 10^4, 10^5, 10^5, 10^5, 10^4, 10^3, 10^5$ \\
\midrule
 global & sparse & grid & 5 & $0, 10^5, 10^2, 10^4, 10^5, 10^5, 10^5, 10^5, 10^5, 10^5, 10^4, 10^4, 10^5, 10^4, 10^5$ \\
\midrule
 global & sparse & grid & 6 & $0, 10^1, 10^0, 10^5, 10^0, 10^5, 10^5, 10^5, 10^2, 10^4, 10^5, 10^5, 10^5, 10^2, 10^5$ \\
\midrule
global & interp. & default & 1--3 & $0, 3\times10^{4},\,\ldots,\,3\times10^{4}$ \\
\midrule
global & interp. & grid & 4 & $0, 10^4, 10^5, 10^5, 10^1, 10^5, 10^5, 10^5, 10^3, 10^4, 10^5, 10^4, 10^4, 10^5, 10^4$ \\
\midrule
global & interp. & grid & 5 & $0, 10^4, 10^2, 10^0, 10^0, 10^5, 10^4, 10^5, 10^4, 10^5, 10^5, 10^5, 10^4, 10^5, 10^5$ \\
\midrule
global & interp. & grid & 6 & $0, 10^4, 10^5, 10^5, 10^5, 10^5, 10^5, 10^5, 10^5, 10^5, 10^5, 10^5, 10^4, 10^3, 10^5$ \\
\bottomrule
local & sparse & default & 1--3 & $0, 3\times10^{4},\,\ldots,\,3\times10^{4}$ \\
\midrule
local & sparse & default & 4 & $0,\,5{\times}10^{5},\,\ldots,\,5{\times}10^{5},\,5{\times}10^{7}$ \\
\midrule
local & sparse & default & 5 & $0,\,5{\times}10^{6},\,\ldots,\,5{\times}10^{6},\,5{\times}10^{7},\,5{\times}10^{7},\,5{\times}10^{7}$ \\
\midrule
local & sparse & default & 6 & $0,\,5{\times}10^{5},\,\ldots,\,5{\times}10^{5}$ \\
\midrule
local & interp. & default & 1--3 & $0, 3\times10^{4},\,\ldots,\,3\times10^{4}$ \\
\midrule
local & interp. & default & 4 & $0,\,5{\times}10^{5},\,\ldots,\,5{\times}10^{5},\,5{\times}10^{7}$ \\
\midrule
local & interp. & default & 5 & $0,\,5{\times}10^{6},\,\ldots,\,5{\times}10^{6},\,5{\times}10^{7},\,5{\times}10^{7},\,5{\times}10^{7}$ \\
\midrule
local & interp. & default & 6 & $0,\,5{\times}10^{5},\,\ldots,\,5{\times}10^{5}$ \\
\bottomrule
\end{tabular}}

\label{tab:lambda_settings_ablations}
\end{table*}
\end{document}

%% file: math_commands.tex
\usepackage{amsmath,amsfonts,bm}

\def\eqref#1{equation~\ref{#1}}

\def\1{\bm{1}}

\DeclareMathAlphabet{\mathsfit}{\encodingdefault}{\sfdefault}{m}{sl}
\SetMathAlphabet{\mathsfit}{bold}{\encodingdefault}{\sfdefault}{bx}{n}

